\documentclass[letterpaper]{article} 
\usepackage{aaai25}  

\makeatletter
\def\maketitle{%
  \par%
  \begingroup 
    \def\thefootnote{\fnsymbol{footnote}}
    \twocolumn[\@maketitle] \@thanks%
  \endgroup%
}
\makeatother

\usepackage{times}  
\usepackage{helvet}  
\usepackage{courier}  
\usepackage[hyphens]{url}  
\usepackage{graphicx} 
\usepackage{natbib}  
\usepackage{caption} 
\usepackage{algorithm}
\usepackage{algorithmic}

\usepackage{newfloat}
\usepackage{listings}
\DeclareCaptionStyle{ruled}{labelfont=normalfont,labelsep=colon,strut=off} 
\floatstyle{ruled}
\newfloat{listing}{tb}{lst}{}
\floatname{listing}{Listing}
\usepackage{multirow}
\usepackage{booktabs}
\usepackage{array}
\usepackage{subcaption}
\usepackage{xcolor}

\usepackage{float}

\usepackage{soul} 
\definecolor{myyellow}{rgb}{1, 0.972549, 0.584313}
\definecolor{mygreen}{rgb}{0.772549, 0.945098, 0.756862}
\definecolor{myblue}{rgb}{0.811764, 0.866666, 0.996078}
\definecolor{myred}{rgb}{0.984313, 0.749019, 0.737254}
\definecolor{darkgreen}{rgb}{0, 0.5001960, 0}
\definecolor{darkred}{rgb}{0.8, 0, 0}
\newcommand{\hlc}[2][yellow]{\sethlcolor{#1}\hl{#2}}

\usepackage{amsmath}
\usepackage{amssymb}

\title{500xCompressor: Generalized Prompt Compression for Large Language Models}
\author {
    Zongqian Li\textsuperscript{\rm 1},
    Yixuan Su\textsuperscript{\rm 1},
    Nigel Collier\textsuperscript{\rm 1}
}
\affiliations {
    \textsuperscript{\rm 1}University of Cambridge\\
    zl510@cam.ac.uk, ys484@cam.ac.uk, nhc30@cam.ac.uk
}

\begin{document}

\maketitle

\begin{abstract}
Prompt compression is crucial for enhancing inference speed, reducing costs, and improving user experience. However, current methods face challenges such as low compression ratios and potential data leakage during evaluation. To address these issues, we propose 500xCompressor, a method that compresses extensive natural language contexts into a minimum of one single special token. The 500xCompressor introduces approximately 0.25\% additional parameters and achieves compression ratios ranging from 6x to 480x. It is designed to compress any text, answer various types of questions, and could be utilized by the original large language model (LLM) without requiring fine-tuning. Initially, 500xCompressor was pretrained on the Arxiv Corpus, followed by fine-tuning on the ArxivQA dataset, and subsequently evaluated on strictly unseen and classical question answering (QA) datasets. The results demonstrate that the LLM retained 62.26-72.89\% of its capabilities compared to using non-compressed prompts. This study also shows that not all the compressed tokens are equally utilized and that K V values have significant advantages over embeddings in preserving information at high compression ratios. The highly compressive nature of natural language prompts, even for fine-grained complex information, suggests promising potential for future applications and further research into developing a new LLM language.
\end{abstract}

\begin{figure}[th!]
    \centering
    \includegraphics[width=0.4\textwidth]{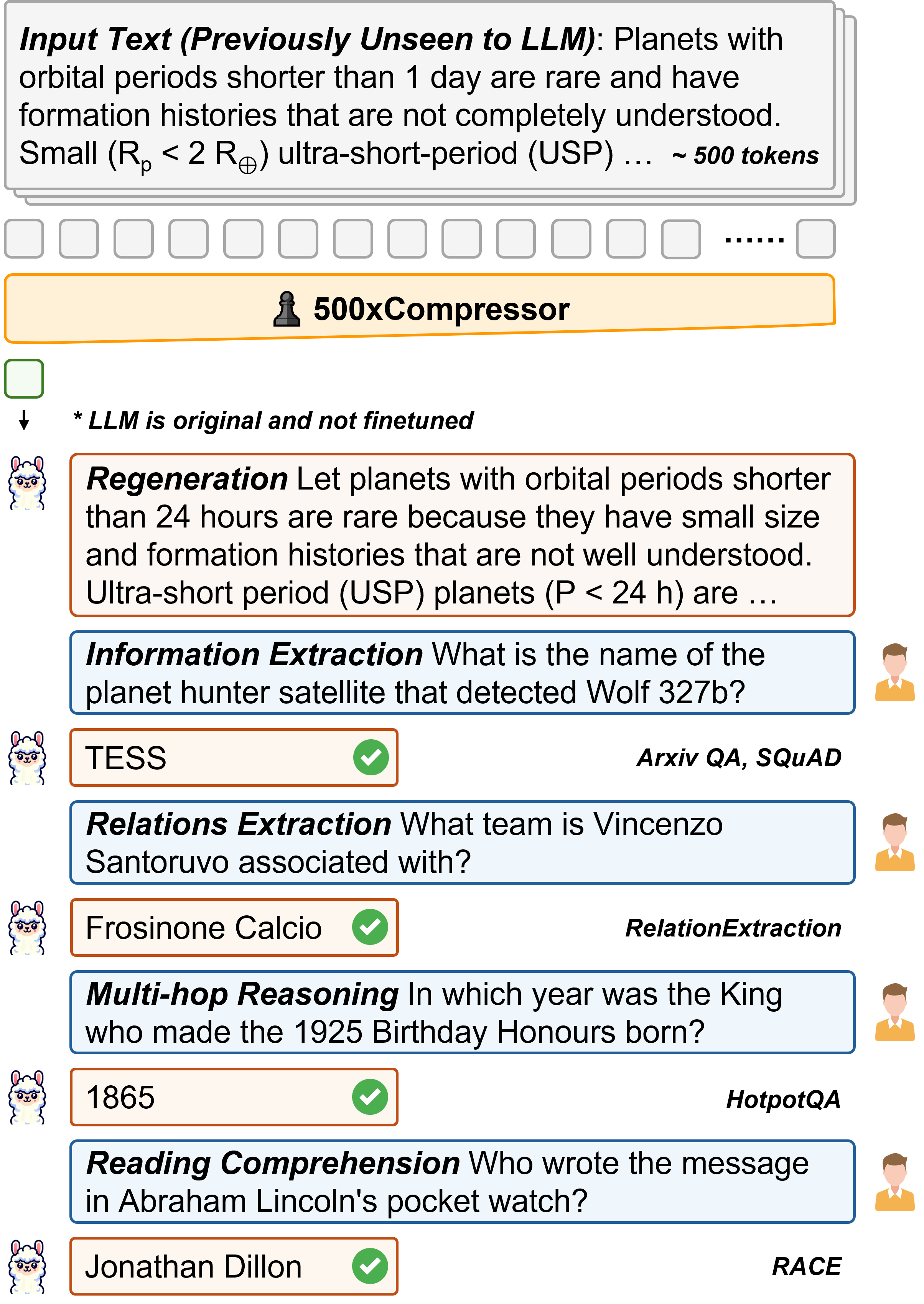}
    \caption{The original text is compressed by 500xCompressor and utilized for downstream tasks.}
    \label{cover_figure}
\end{figure}

\begin{figure*}[t]
    \centering
    \includegraphics[width=0.95\textwidth]{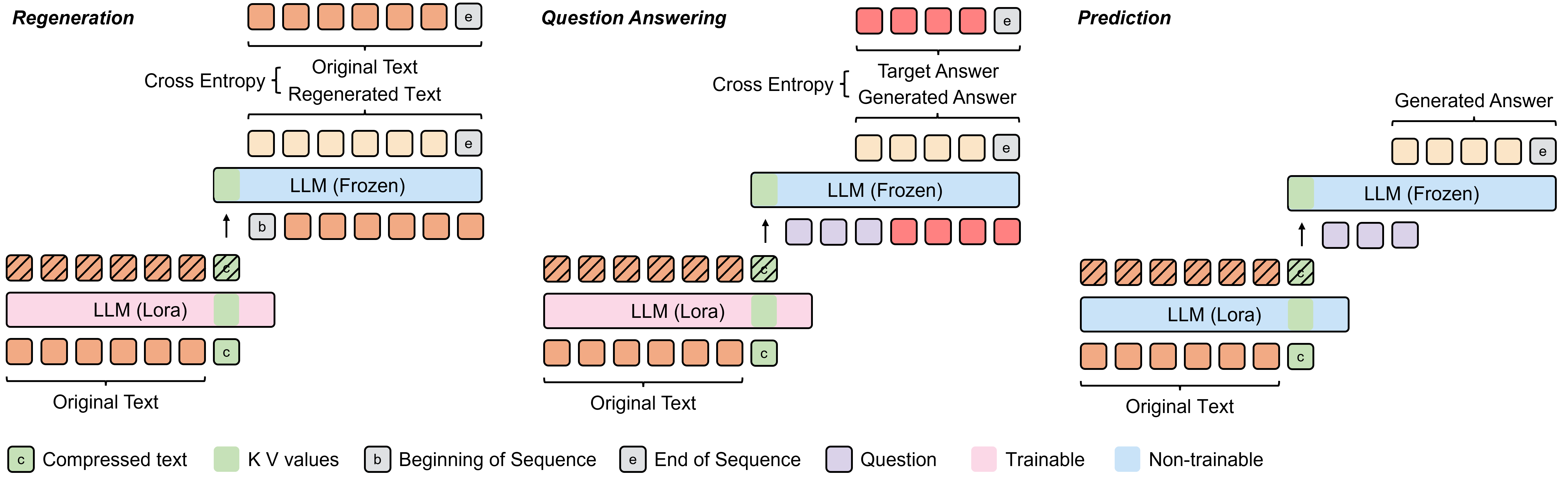}
    \caption{Process of pretraining (left), fine-tuning (middle), and prediction (right) with 500xCompressor.}
    \label{train}
\end{figure*}

\section{Introduction}

Long prompts present several significant challenges in natural language processing applications, including decreased inference speed, increased computation cost, and a negative impact on user experience. Additionally, the context length limit restricts model performance and application scenarios, creating a strong demand for reducing prompt length.

Two primary methods for prompt compression have been proposed: hard prompt and soft prompt. Hard prompt methods, such as SelectiveSentence \cite{li-etal-2023-compressing} and LLMLingua \cite{jiang-etal-2023-llmlingua}, eliminate low-information sentences, words, or even tokens. In contrast, soft prompt methods, including GIST \cite{mu2024learning}, AutoCompressor \cite{chevalier2023adapting}, and ICAE \cite{ge2024incontext}, compress natural language tokens into a small number of special tokens. However, these methods have problems such as low compression ratios, unclear information loss, and potential data leakage during evaluation, as discussed in detail in Section \ref{related-work}. For instance, ICAE achieves compression ratios no higher than 15x, and the win rate evaluation metric fails to quantitatively capture the extent of information loss during compression. Additionally, evaluation texts sourced from the Pile dataset might overlap with the training data for LlaMa series models, raising concerns that the generated content could be retrieved from the memory of the LLM rather than the compressed tokens.

To solve these problems, we propose 500xCompressor, illustrated in Figure \ref{cover_figure}. This method compresses prompts of approximately 500 tokens into a minimum of one token, allowing the compressed tokens to regenerate the original texts or be used for QA. Although trained on the Arxiv Corpus and ArxivQA dataset, 500xCompressor could generalize to answer other types of questions. Drill down analysis demonstrates that complex information, such as proper nouns, special names, and numbers, could be accurately compressed and retrieved.

500xCompressor retains the advantages of previous methods and introduces several additional features. Similar to previous soft prompt methods, 500xCompressor is generalized and non-selective, capable of compressing unseen texts across various topics for QA, demonstrating its generalization ability. Unlike selective compression methods, 500xCompressor aims to regenerate the entire original text, ensuring that all tokens in the original text contribute to the compressed tokens. Moreover, the compressed prompts could be used to regenerate original texts or for QA without requiring fine-tuning of the LLM, preserving the LLM's original capabilities and enhancing the convenience of using compressed tokens.

In addition to these existing advantages, we provide contributions in three key areas:

\begin{itemize}
    \item High Compression Ratio: This study evaluates the compression model with one, four, and sixteen tokens to compress up to 500 tokens, achieving compression ratios from 6x to 480x. These ratios significantly surpass previous research, which reported ratios of less than 50x, fully exploring the upper limit of prompt compression.
    
    \item Strict Unseen Evaluation Set: The evaluation texts in the Arxiv Corpus and ArxivQA dataset were published after January 2024, which are new, domain-specific texts not used for training the original LLM or the compression model. Thus, when compressed texts are input to the LLM, the outputs primarily derive from the compressed texts, not from the pre-existing knowledge of the LLM.
    
    \item Quantitative Analysis of Information Loss: The compressed texts are used for extractive QA, where the answer to a question is a context span, allowing specific target answers. This setup enables a quantitative comparison of 500xCompressor with baseline and gold standards, providing a detailed analysis of information loss during prompt compression.
\end{itemize}

In this paper, the architecture and mechanism of 500xCompressor are first introduced in Section \ref{Methods}, including how to train and use the compression model. After that, Section \ref{Experiments} explains the training and evaluation datasets, the baseline, and the evaluation metrics. The evaluation results for regeneration and QA are presented in Section \ref{Results}, with ablation studies analyzing the factors influencing the compression models. This is followed by discussions in Section \ref{Discussions}, and finally, the sections on related work and conclusions.

\section{Methods}
\label {Methods}

\subsection{Training}

The training process for the compression model is illustrated in Figure \ref{train}, encompassing both pretraining and fine-tuning phases. The compression model comprises two components: an encoder and a decoder, functioning similarly to an autoencoder and comparable to ICAE. The encoder is the frozen LLM $\mathbf{\Theta_{\text{LLM}}}$ with trainable LoRA parameters $\mathbf{\Theta_{\text{Lora}}}$, while the decoder is the original frozen LLM $\mathbf{\Theta_{\text{LLM}}}$. The encoder receives the original text tokens $\mathbf{T} = (t_1, t_2, \ldots, t_l)$ and compressed tokens $\mathbf{C} = (c_1, c_2, \ldots, c_k)$. Through attention mechanisms, the information in the text is encoded into the compressed tokens, whose K V values in each layer of the LLM $\mathbf{H_C}$ are output and passed to the decoder.

During pretraining, the inputs of the decoder are the K V values of the compressed tokens from the encoder, the beginning of sequence token, and the original text tokens $(\mathbf{H_C}, \mathbf{[BOS]}, \mathbf{T})$. Teacher forcing is employed to guide the LLM in regenerating the original text based on the K V values, using the end of sequence token $\mathbf{[EOS]}$ to signal when to stop. The cross-entropy loss between the output of the deocder and the original text is calculated and used to train the LoRA parameters in the encoder via backpropagation:

{
\small
\begin{equation}
\mathcal{L}_{\text{P}} = - \sum_{i=1}^{l} \log P(t_i | \mathbf{H_C}, \mathbf{[BOS]}, t_{1:i-1}; \mathbf{\Theta_{\text{LLM}}}, \mathbf{\Theta_{\text{Lora}}})
\end{equation}
}

For instruction fine-tuning, the process mirrors pretraining. However, instead of the original texts, the decoder is provided with questions $\mathbf{Q} = (q_1, q_2, \ldots, q_m)$ and answers $\mathbf{A} = (a_1, a_2, \ldots, a_n)$, which are used to train the LLM to retrieve information from the K V values of the compressed tokens and generate answers:

{
\small
\begin{equation}
\mathcal{L}_{\text{F}} = - \sum_{j=1}^{n} \log P(a_j | \mathbf{H_C}, q_{1:m}, a_{1:j-1}; \mathbf{\Theta_{\text{LLM}}}, \mathbf{\Theta_{\text{Lora}}})
\end{equation}
}

The training process ensures no data leakage, as the original LLM parameters in both the encoder and decoder remain unchanged, and no additional parameters are introduced in the decoder. Thus, no information is saved in the decoder.

A key difference between 500xCompressor and ICAE is the input of the decoder. The decoder of ICAE input is the embeddings for the compressed tokens, whereas 500xCompressor uses the K V values for the compressed tokens. K V values could encapsulate more information, do not increase inference time, and have small impact on GPU memory usage. In addition, this paper uses the $\mathbf{[BOS]}$ token to trigger the LLM to regenerate the compressed texts, while ICAE creates a trainable new token.

\subsection{Prediction}

During prediction, all encoder and decoder parameters are frozen. The original text is fed into the encoder, which saves the information into compressed tokens via attention mechanisms. These compressed tokens' K V values are then input into the decoder, which regenerates the compressed text when triggered by the $\mathbf{[BOS]}$ token or generates an answer based on a given question:

{
\small
\begin{equation}
\hat{t}_i = \arg\max_{t_i} P(t_i | \mathbf{H_C}, \mathbf{[BOS]}, t_{1:i-1}; \mathbf{\Theta_{\text{LLM}}})
\end{equation}
}

{
\small
\begin{equation}
\hat{a}_j = \arg\max_{a_j} P(a_j | \mathbf{H_C}, q_{1:m}, a_{1:j-1}; \mathbf{\Theta_{\text{LLM}}})
\end{equation}
}

By replacing the original text tokens with compressed tokens, the speed of answering questions is increased. This is because, in inference, each token in the question or generated answer must attend to the previous tokens. Replacing a large number of original text tokens with a small number of compressed tokens reduces the computational load.

\begin{figure*}[t]
\centering
\begin{subfigure}[b]{0.245\textwidth}
    \centering
    \includegraphics[width=\textwidth]{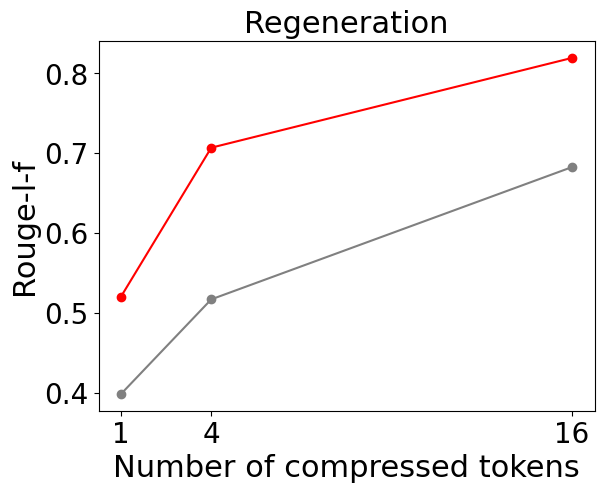}
    \label{reg-1}
\end{subfigure}
\begin{subfigure}[b]{0.245\textwidth}
    \centering
    \includegraphics[width=\textwidth]{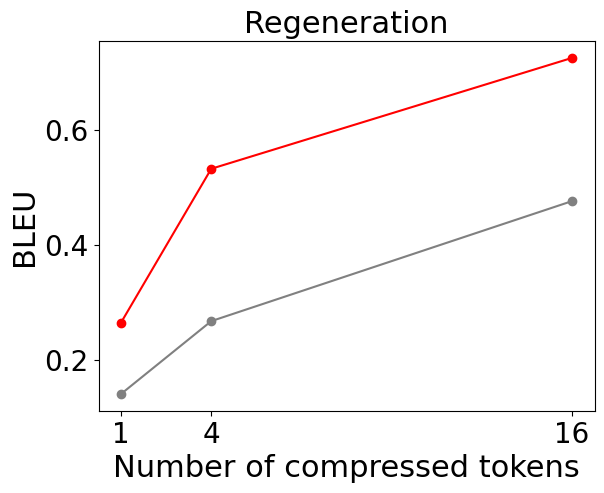}
    \label{reg-2}
\end{subfigure}
\begin{subfigure}[b]{0.245\textwidth}
    \centering
    \includegraphics[width=\textwidth]{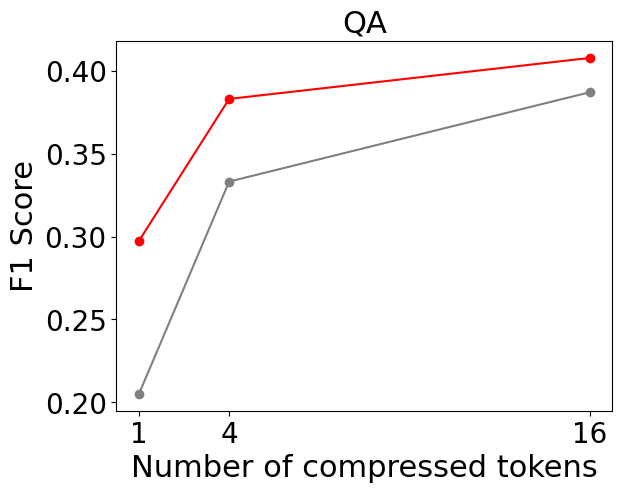}
    \label{qa-1}
\end{subfigure}
\begin{subfigure}[b]{0.245\textwidth}
    \centering
    \includegraphics[width=\textwidth]{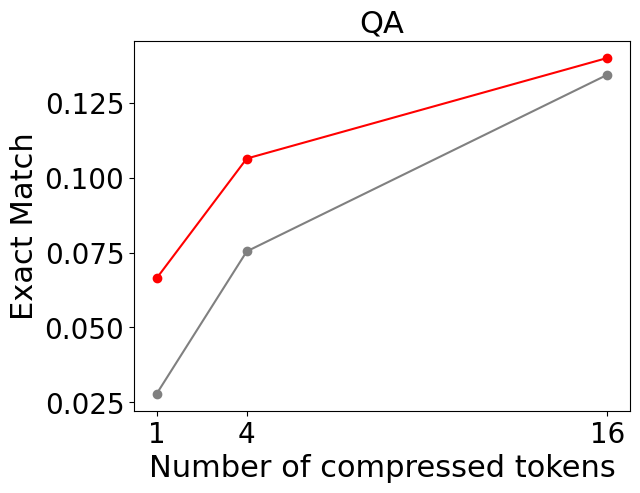}
    \label{qa-2}
\end{subfigure}
\\
\vspace{-1.5em}
\begin{subfigure}[b]{0.3\textwidth}
    \centering
    \includegraphics[width=\textwidth]{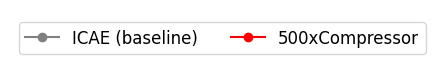}
    \label{reg-qa-legend-1}
\end{subfigure}
\vspace{-1.5em}
\caption{Evaluation results for text regeneration on the Arxiv Corpus.}
\label{reg-qa}
\end{figure*}

\section{Experiments}
\label{Experiments}

\subsection{Datasets}

The Arxiv Corpus was used to pretrain 500xCompressor, and the compression model was then fine-tuned on the ArxivQA dataset. After that, five benchmarks were used to evaluate the compression models for various abilities: ArxivQA and SQuAD \cite{rajpurkar2016squad} for information extraction, RelationExtraction \cite{levy2017zero} for relation extraction, HotpotQA \cite{yang2018hotpotqa} for multi-hop reasoning, and RACE \cite{lai2017race} for reading comprehension. Among these datasets, ArxivQA is introduced in this paper, while the others are classical QA datasets.

The Arxiv Corpus is a collection of abstracts from Arxiv papers published before April 2024, used to pretrain the compression model. Papers published before July 2023 constitute the training set, while papers published after January 2024 form the development and test sets. In the test set, abstracts with lengths of at least 96, 192, 288, 384, and 480 tokens are selected and grouped to test the regeneration performance of the LLMs.

There are several reasons for creating the Arxiv Corpus for pretraining the compression model. Arxiv abstracts are high-quality plain texts rich in expert knowledge. The upload date for each paper is clear, making it easy to determine when the text was made public and whether it is in the training corpus of the LLM. The knowledge cutoff for the LLaMa-3 series models is March 2023, hence the test set is strictly unseen by the LLM. This confirms that the regenerated content is in the compressed tokens, not in the "memory" or parameters of the original LLM. Additionally, the Arxiv abstracts are officially uploaded to Kaggle by Cornell University, making sure that they are easy to access and there is no authority issues. In contrast, the Pile dataset contains numerous unauthorized articles, making it unsuitable for training the compression model.

The ArxivQA dataset is generated from the Arxiv Corpus by LLaMa-3-70b-chat. Five extractive QA pairs are created for each 96-token abstract, with the number of QA pairs increasing proportionally with the length of the abstract. Each data record contains an abstract as the context, a question based on this context, and an answer to the question. The QA pairs in the training and development sets are created from abstracts in the training set, while those in the test set are based on abstracts in the test set.

The ArxivQA dataset has several advantages. All contexts in its test set are strictly unseen by the LLM, preventing potential data leakage. The dataset consists of extractive QA pairs with standard answers extracted from the context, facilitating easy calculation of information loss by comparing generated answers with standard answers. Due to the nature of Arxiv abstracts, the questions in the ArxivQA dataset contain more domain-specific knowledge and are more challenging. Besides, the powerful capabilities of LLaMa-3-70b-chat guarantee the quality of the QA pairs and generate questions on a wide variety of topics.

\subsection{Baseline and Gold Standards}

ICAE was used to compare with 500xCompressor for the task of extractive QA. ICAE is a soft prompt method that compresses natural language tokens into a small number of special tokens, retaining only the embeddings of these compressed tokens. After training the compression model, the original LLM does not need to be fine-tuned for decoding and using the compressed tokens in ICAE. Considering the soft prompt methods and no fine-tuning for decoding, ICAE was chosen as the baseline.

Two gold standards were used: zero-shot full context and instruct full context. For zero-shot full context, the entire text was given to the LLM without instruction to guide the extractive QA tasks. For instruct full context, the instructions were provided to the LLM.

\subsection{Evaluation Metrices}

Rouge-l-f (Recall-Oriented Understudy for Gisting Evaluation) and BLEU (Bilingual Evaluation Understudy) scores were used to evaluate the difference between the regenerated text and the original text. Rouge-l-f focuses on the longest common subsequence between the two texts, capturing both recall and precision aspects. This metric balances the need for both accurately recalling relevant information and ensuring the precision of the regenerated content. On the other hand, the BLEU score measures the precision of n-grams in the regenerated text against the original text, making it effective for assessing the fluency and accuracy of text generation. By utilizing both Rouge-l-f and BLEU, we achieve a comprehensive assessment, where Rouge-l-f ensures a balanced evaluation of recall and precision, and BLEU emphasizes precision in textual similarity.

F1 score and EM (Exact Match) were used to evaluate the results for the extractive QA benchmarks. The F1 score, which is the harmonic mean of precision and recall, provides a balanced measure of the accuracy of the model in identifying the correct answers. EM is a stricter metric that assesses whether the predicted answer matches the ground truth answer exactly. These metrics are crucial for extractive QA tasks as they ensure that the model not only identifies relevant information accurately (high F1 score) but also captures the precise answer required (high EM). Their combined use offers a detailed evaluation of model performance, balancing overall accuracy with exactness.

\section{Results}
\label{Results}

\subsection{Text Regeneration}

Figure \ref{reg-qa} (left) presents the results for LLaMa-3-8b-chat in regenerating the original Arxiv abstracts from the compressed tokens. Compressed tokens numbering 1, 4, and 16 were used to compress 96, 192, 288, 384, and 480 tokens, achieving compression ratios from 6x to 480x. For each number of compressed tokens, two points represent the average evaluation results for the regenerated texts by ICAE (baseline) and 500xCompressor (ours). Higher Rouge-l-f or BLEU Scores indicate greater similarity between the original text and the regenerated text.

The comparison between the red and gray lines shows that 500xCompressor outperforms ICAE in regenerating compressed texts. The Rouge-l-f for 500xCompressor exceed those for ICAE at all compression ratios, with differences ranging from 12.18\% to 18.96\%. Similarly, the BLEU score differences range from 12.41\% to 26.50\%. Accurate text regeneration is crucial for downstream tasks such as QA.

When the number of compressed tokens is small, the quality of the regenerated text decreases rapidly. The slopes of the red and gray lines are similar, but the slopes between 1-4 tokens are steeper than those between 4-16 tokens. This shows that the rates of decline for Rouge-l-f and BLEU Scores increase with higher compression ratios for both 500xCompressor and ICAE.

Examples of regenerated texts by 500xCompressor and ICAE are shown in Table \ref{example-table} (upper). Compared to ICAE, the text regenerated by 500xCompressor is more similar to the original text, with fewer mistakes, less paraphrasing, no information loss and hallucination in this example.

\begin{table*}[ht]
\centering
\begin{tabular}{p{0.3\textwidth} p{0.3\textwidth} p{0.3\textwidth}}
\hline
\multicolumn{1}{c}{\textbf{Original}} & \multicolumn{1}{c}{\textbf{500xCompressor (ours)}} & \multicolumn{1}{c}{\textbf{ICAE (baseline)}} \\
\hline
\multicolumn{1}{p{0.3\textwidth}}{We show that every reciprocity sheaf gives rise to a cycle (pre)module in the sense of Rost over a perfect field. Over a perfect field of positive characteristic, we show that the first cohomology group of a logarithmic de Rham-Witt sheaf has a partial cycle module structure. As a consequence, we show that Kato complexes of logarithmic de Rham-Witt sheaves satisfy functoriality properties similar to Rost's cycle complexes. \vspace{0.5em}} & 
\multicolumn{1}{p{0.3\textwidth}}{We show that every reciprocity sheaf gives rise to a cycle (pre)module in the sense of Rost over a perfect field. Over a perfect field of positive characteristic, we show that the first cohomology group of a logarithmic de Rham-Witt \hlc[myred]{cycle} module has a partial cycle structure. As a consequence, we show that Kato \hlc[myred]{modules} of logarithmic de Rham-Witt \hlc[myred]{complexes} satisfy \hlc[mygreen]{functorial} properties similar to Rost's cycle complexes. \vspace{0.5em}} & 
\multicolumn{1}{p{0.3\textwidth}}{We show that every \hlc[myred]{sheaf reciprocity} gives rise to a cycle (pre)module \hlc[myyellow]{\textit{in the sense of Rost}} over a \hlc[myyellow]{\textit{perfect field}} \hlc[myblue]{Rost cycle}. \hlc[mygreen]{In the perfect field case}, we show that \hlc[mygreen]{over a positive characteristic field}, the first logarithmic de Rham\hlc[myyellow]{\textit{-Witt sheaf}} cohomology group of a \hlc[myred]{Witt log-Witt cycle} has a partial \hlc[myyellow]{\textit{cycle module structure}} \hlc[myblue]{decomposition}. As a consequence, we show that Kato's \hlc[myblue]{cycle} complexes \hlc[myyellow]{\textit{of logarithmic de Rham-Witt sheaves}} satisfy functoriality properties similar to Rost cycle complexes. \vspace{0.5em}} \\
\hline
\multicolumn{3}{p{0.9\textwidth}}{\textbf{Q:} What type of sheaf gives rise to a cycle premodule?} \vspace{0.25em} \\
\multicolumn{1}{p{0.3\textwidth}}{\textbf{A:} Every reciprocity sheaf.} \vspace{0.25em} & 
\multicolumn{1}{p{0.3\textwidth}}{\textbf{A:} a reciprocity sheaf} \vspace{0.25em} & 
\multicolumn{1}{p{0.3\textwidth}}{\textbf{A:} a \hlc[myyellow]{\textit{reciprocity}} sheaf of \hlc[myblue]{(logarithmic) differential forms}} \vspace{0.25em}\\
\multicolumn{3}{p{0.9\textwidth}}{\textbf{Q:} Over what type of field do we show that Kato complexes satisfy functoriality properties?} \vspace{0.25em} \\
\multicolumn{1}{p{0.3\textwidth}}{\textbf{A:} Over a perfect field of positive characteristic.} \vspace{0.25em} & 
\multicolumn{1}{p{0.3\textwidth}}{\textbf{A:} perfect fields of positive characteristic} \vspace{0.25em} & 
\multicolumn{1}{p{0.3\textwidth}}{\textbf{A:} a perfect field of \hlc[myyellow]{\textit{positive}} characteristic \hlc[myblue]{zero}} \vspace{0.25em} \\
\multicolumn{3}{p{0.9\textwidth}}{\textbf{Q:} What is the structure of the first cohomology group of a logarithmic de Rham-Witt sheaf?} \vspace{0.25em} \\
\multicolumn{1}{p{0.3\textwidth}}{\textbf{A:} a partial cycle module structure} \vspace{0.25em} & 
\multicolumn{1}{p{0.3\textwidth}}{\textbf{A:} a partial cycle \hlc[myred]{complex}} \vspace{0.25em} & 
\multicolumn{1}{p{0.3\textwidth}}{\textbf{A:} a partial \hlc[myblue]{Kato} cycle \hlc[myred]{complex}} \vspace{0.25em} \\
\multicolumn{3}{p{0.9\textwidth}}{\textbf{Q:} What type of complexes satisfy functoriality properties similar to Rost's cycle complexes?} \vspace{0.25em} \\
\multicolumn{1}{p{0.3\textwidth}}{\textbf{A:} Kato complexes of logarithmic de Rham-Witt sheaves} \vspace{0.25em} & 
\multicolumn{1}{p{0.3\textwidth}}{\textbf{A:} Kato-Witt \hlc[myblue]{cycle} complexes \hlc[myyellow]{\textit{of logarithmic de Rham-Witt sheaves}}} \vspace{0.25em} & 
\multicolumn{1}{p{0.3\textwidth}}{\textbf{A:} Kato's complexes \hlc[myyellow]{\textit{of logarithmic de Rham-Witt sheaves}}} \vspace{0.25em} \\
\hline
\end{tabular}
\caption{Examples of regenerated texts and QA pairs provided by 500xCompressor and ICAE. 96 tokens of the original text were compressed into 4 tokens, which were then used for QA. Differences between the gold standard and the output include mistakes (\hlc[myred]{red}, containing incorrect text), information loss (\hlc[myyellow]{\textit{yellow and italic}}, missing some text), hallucinations (\hlc[myblue]{blue}, including text not present in the original), and paraphrasing (\hlc[mygreen]{green}, rephrasing the original text).}
\label{example-table}
\end{table*}

\subsection{Question Answering}

Figure \ref{reg-qa} (right) shows the results for LLaMa-3-8b-chat in answering extractive ArxivQA questions based on the compressed texts. For both 500xCompressor and ICAE, original contexts with 96-480 tokens were compressed into 1, 4, or 16 tokens, and the LLM performed QA based solely on these compressed tokens. Higher F1 scores or EM values indicate more accurate answers.

500xCompressor outperforms ICAE on the ArxivQA dataset, showing improvements of 2.06-9.23\% in F1 score and 0.56-7.20\% in EM. As the number of compressed tokens decreases, the rates of decline for F1 score and EM are higher for ICAE than for 500xCompressor. The higher the compression ratio is, the higher the decreasing speed is. This demonstrates that ICAE loses more information and tends to degrade faster with higher compression ratios.

QA examples are shown in Table \ref{example-table} (lower). Consistent with text regeneration, answers provided by ICAE lose more information and contain more hallucinations compared to those provided by 500xCompressor. Although the quality of the given answers is related to the quality of the regenerated text, errors in the regenerated text may not always appear in the QA responses and vice versa. Even if the regenerated text contains some errors, these errors may not affect downstream tasks such as QA. For example, 500xCompressor regenerates "of a logarithmic de Rham-Witt" but omit this information in its answer. Conversely, even if 500xCompressor incorrectly regenerates "Kato modules" instead of "Kato complexes", it could still produce the correct term "complexes" in its answer.

To validate the generalization ability of the compression models on additional tasks, both 500xCompressor and ICAE were tested on four more benchmarks, each focusing on different capabilities. The results are presented in Table \ref{benchmarks} and Figure \ref{benchmarks-figures}. Overall, 500xCompressor outperforms ICAE across various tasks. For information extraction, higher compression ratios accentuate the advantage of 500xCompressor. However, this trend does not consistently hold for other benchmarks. When the compression ratio is high, ICAE degrades more quickly than 500xCompressor, making the latter’s improvements more pronounced. For the 500$\rightarrow$1 compression, 500xCompressor surpasses ICAE across all benchmarks, with the most significant improvement being an 18.62\% increase in F1 score for relation extraction. Notably, the performance of 500xCompressor even improves in HotpotQA and RACE from 500$\rightarrow$4 to 500$\rightarrow$1, despite the quadrupled compression ratio.

\begin{table*}[t]
    \centering
    \begin{tabular}{ccccccccccccc}
        \hline
        {} & \multicolumn{2}{c}{\textbf{ArxivQA}} & \multicolumn{2}{c}{\textbf{SQuAD}} & \multicolumn{2}{c}{\textbf{RE}} & \multicolumn{2}{c}{\textbf{HotpotQA}} & \multicolumn{2}{c}{\textbf{RACE}} & \multicolumn{2}{c}{\textbf{Average}} \\
        {} & \multicolumn{4}{c}{Information Extraction} & \multicolumn{2}{c}{Relationships} & \multicolumn{2}{c}{Multi-hop} & \multicolumn{2}{c}{Comprehension} & \multicolumn{2}{c}{} \\
        \cline{2-13}
         & F1 & EM & F1 & EM & F1 & EM & F1 & EM & F1 & EM & F1 & EM \\
        \hline
        Zero-shot & 28.41 & 0.04 & 29.95 & 1.02 & 25.75 & 1.39 & 29.32 & 3.42 & 15.31 & 0.59 & 25.75 & 1.29 \\
        Instruct & 55.90 & 13.74 & 70.65 & 36.79 & 71.36 & 53.15 & 69.37 & 43.19 & 39.52 & 13.94 & 61.36 & 32.16 \\
        \hline
        Ours (500$\rightarrow$16) & 40.77 & 14.00 & 50.01 & 28.59 & 68.73 & 50.44 & 41.68 & 22.99 & 35.42 & 10.97 & 45.32 & 25.40 \\
        ICAE (500$\rightarrow$16) & 38.70 & 13.44 & 51.94 & 30.37 & 65.94 & 44.64 & 42.11 & 24.57 & 23.69 & 9.49 & 44.48 & 24.50 \\
        $\Delta$ & \textcolor{darkgreen}{2.06} & \textcolor{darkgreen}{0.56} & \textcolor{darkred}{-1.92} & \textcolor{darkred}{-1.78} & \textcolor{darkgreen}{2.79} & \textcolor{darkgreen}{5.80} & \textcolor{darkred}{-0.42} & \textcolor{darkred}{-1.57} & \textcolor{darkgreen}{1.73} & \textcolor{darkgreen}{1.48} & \textcolor{darkgreen}{0.84} & \textcolor{darkgreen}{0.89} \\
        \hline
        Ours (500$\rightarrow$4) & 38.30 & 10.64 & 49.66 & 27.54 & 63.72 & 45.69 & 36.86 & 19.69 & 21.49 & 7.86 & 42.00 & 22.28 \\
        ICAE (500$\rightarrow$4) & 33.31 & 7.54 & 47.48 & 26.49 & 67.28 & 50.91 & 40.39 & 23.75 & 20.06 & 6.82 & 41.70 & 23.10 \\
        $\Delta$ & \textcolor{darkgreen}{4.98} & \textcolor{darkgreen}{3.10} & \textcolor{darkgreen}{2.17} & \textcolor{darkgreen}{1.05} & \textcolor{darkred}{-3.56} & \textcolor{darkred}{-5.22} & \textcolor{darkred}{-3.53} & \textcolor{darkred}{-4.06} & \textcolor{darkgreen}{1.43} & \textcolor{darkgreen}{1.03} & \textcolor{darkgreen}{0.30} & \textcolor{darkred}{-0.81} \\
        \hline
        Ours (500$\rightarrow$1) & 29.73 & 10.60 & 42.86 & 22.75 & 63.09 & 45.31 & 37.47 & 21.13 & 21.75 & 7.56 & 38.98 & 21.47 \\
        ICAE (500$\rightarrow$1) & 20.50 & 3.40 & 27.20 & 10.66 & 44.46 & 27.10 & 22.44 & 9.38 & 13.71 & 4.16 & 25.66 & 10.94 \\
        $\Delta$ & \textcolor{darkgreen}{9.23} & \textcolor{darkgreen}{7.20} & \textcolor{darkgreen}{15.66} & \textcolor{darkgreen}{12.09} & \textcolor{darkgreen}{18.62} & \textcolor{darkgreen}{18.21} & \textcolor{darkgreen}{15.02} & \textcolor{darkgreen}{11.74} & \textcolor{darkgreen}{8.03} & \textcolor{darkgreen}{3.40} & \textcolor{darkgreen}{13.31} & \textcolor{darkgreen}{10.53} \\
        \hline
    \end{tabular}
    \caption{Evaluation results across five benchmarks focusing on various capabilities. The first three lines represent gold standards. $\Delta$ shows the difference between 500xCompressor (ours) and ICAE (baseline). \textcolor{darkgreen}{Green} indicates that 500xCompressor outperforms ICAE, while \textcolor{darkred}{red} indicates that the baseline surpasses our method.}
    \label{benchmarks}
\end{table*}

\begin{figure*}[ht!]
\centering
\begin{subfigure}[b]{0.25\textwidth}
    \centering
    \includegraphics[width=\textwidth]{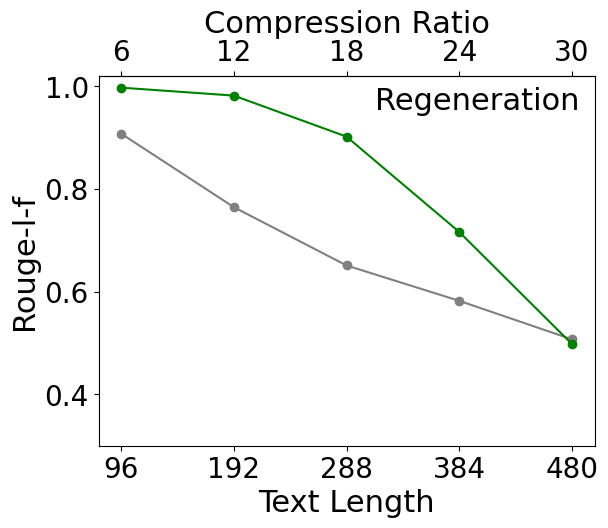}
    \label{reg-rouge-1}
\end{subfigure}
\begin{subfigure}[b]{0.25\textwidth}
    \centering
    \includegraphics[width=\textwidth]{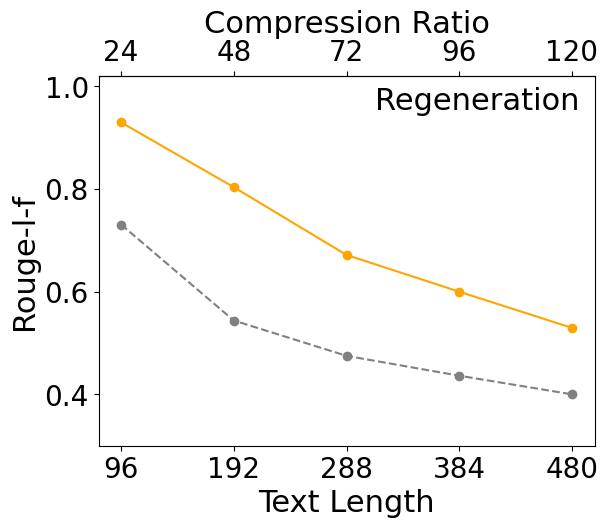}
    \label{reg-rouge-2}
\end{subfigure}
\begin{subfigure}[b]{0.25\textwidth}
    \centering
    \includegraphics[width=\textwidth]{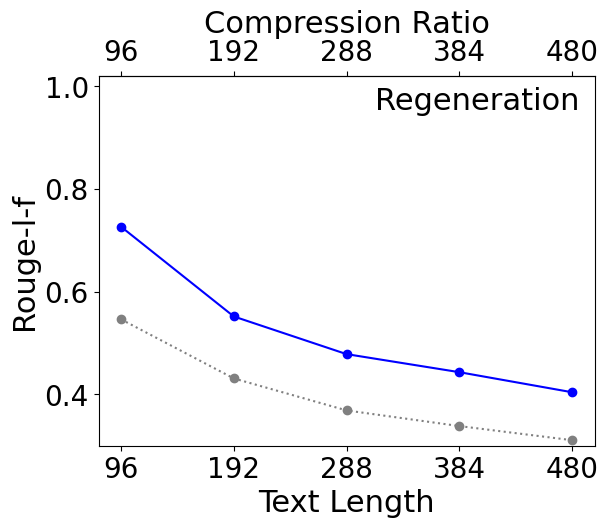}
    \label{reg-rouge-3}
\end{subfigure}
\\
\begin{subfigure}[b]{0.25\textwidth}
    \centering
    \includegraphics[width=\textwidth]{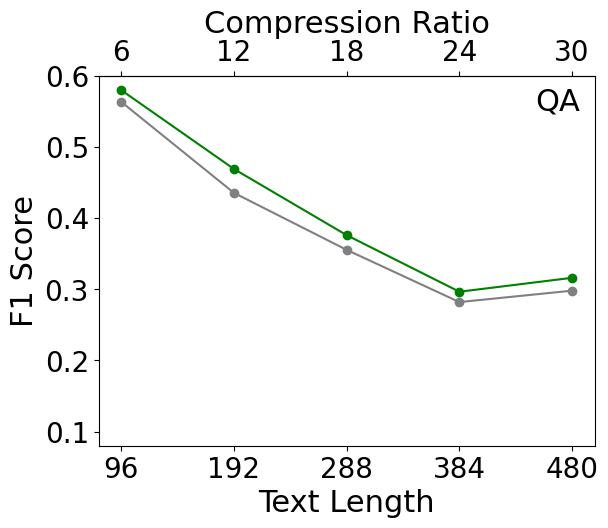}
    \label{qa-f1-1}
\end{subfigure}
\begin{subfigure}[b]{0.25\textwidth}
    \centering
    \includegraphics[width=\textwidth]{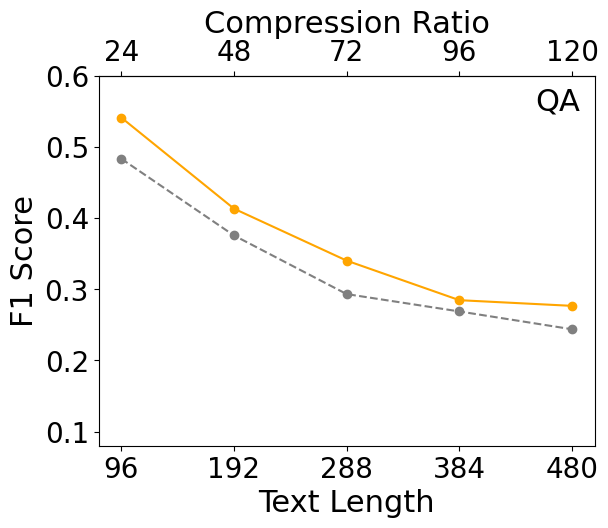}
    \label{qa-f1-2}
\end{subfigure}
\begin{subfigure}[b]{0.25\textwidth}
    \centering
    \includegraphics[width=\textwidth]{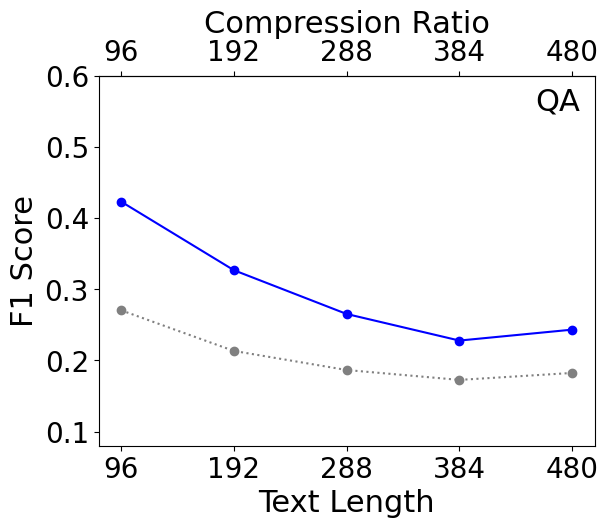}
    \label{qa-f1-3}
\end{subfigure}
\\
\vspace{-0.5em}
\begin{subfigure}[b]{0.5\textwidth}
    \centering
    \includegraphics[width=\textwidth]{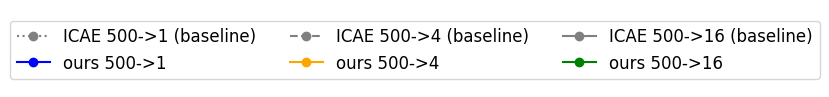}
    \label{ablation-legend}
\end{subfigure}
\vspace{-1.5em}
\caption{Evaluation results for text regeneration on the Arxiv Corpus and for QA on the ArxivQA dataset.}
\label{ablation}
\end{figure*}

\subsection{Ablation Studies}

The performance of compression models is influenced by several factors, including the compression method (ICAE or 500xCompressor), task type (benchmark), context length (length of text to be compressed), and compression ratio (number of compressed tokens). In Figure \ref{ablation}, the X-axis represents context length or compression ratio, and the Y-axis shows evaluated results for different 500xCompressor and ICAE models.

For regeneration, 500xCompressor does not equally utilize all compressed tokens. When the number of compressed tokens decreases from 16 to 4, the regenerated texts by 500xCompressor exhibit similar Rouge-l-f scores. Further reduction to 1 token results in noticeable differences in Rouge-l-f scores. This trend is not observed for ICAE regeneration but is consistent for QA with 500xCompressor. Thus, 500xCompressor is more efficiently to use a small number of compressed tokens.

For QA, the improvement of 500xCompressor is pronounced at higher compression ratios. 500xCompressor and ICAE perform similarly for 500$\rightarrow$16, but the difference increases for 500$\rightarrow$4 and significantly for 500$\rightarrow$1. 500xCompressor keeps more information than ICAE when the number of compressed tokens is small and the compression ratio is high.

\section{Discussions}
\label{Discussions}

The differences between 500xCompressor and ICAE could be better understood by comparing them to Prompt Tuning \cite{lester-etal-2021-power} and Prefix Tuning \cite{li-liang-2021-prefix}. In Prompt Tuning, prefixed special tokens are trained to guide the model in completing specific downstream tasks. Similarly, ICAE compresses contexts into prefixed special tokens for downstream tasks. Unlike Prompt Tuning, Prefix Tuning also trains the K V values associated with the prefixed special tokens. 500xCompressor, akin to Prefix Tuning, compresses texts into the K V values of prefixed special tokens. In Prompt Tuning or Prefix Tuning, the prefixed special tokens (and their K V values) only encode the instruction for the downstream task. However, in ICAE and 500xCompressor, these tokens compress fine-grained information within the context in addition to the instruction.

There are three ways to understand the compressed tokens generated from natural language tokens: as memory slots, a new modality, and a new LLM language. Ge et al. associated text compression in representation learning with working memory in cognitive science, viewing compressed tokens as an efficient way for LLMs to store knowledge, aligning with the idea that "language modeling is compression" \cite{ge2024incontext}. Inspired by multimodality, Cheng et al. interpreted text compression as modality fusion, where compressed tokens, combined with natural language tokens, provide more information and have higher information density \cite{cheng2024xrag}. Jiang et al. treated the compressed prompt as a new language for LLM \cite{jiang-etal-2023-llmlingua}. There are three elements that define a language: encoding information, transmitting information, and adaptive evaluation. The compressed tokens could regenerate the original text, indicating that the information has been effectively encoded. Furthermore, these tokens can be used for downstream tasks and answer related questions, demonstrating their ability to transfer information. The ability of the compression models to handle unseen texts further shows their generalization ability and adaptability. These features make compressed tokens an efficient new language for LLMs.

500xCompressor shows better scalability than ICAE. As shown in Table \ref{benchmarks}, the performance of ICAE decreases faster than 500xCompressor as the number of compressed tokens decreases. Figure \ref{ablation} further illustrates that the improvement of 500xCompressor is more pronounced with higher compression ratios. These results indicate that the performance of 500xCompressor is less affected by high compression ratios or large text lengths compared to ICAE.

Both 500xCompressor and ICAE exhibit similar improvements in inference speed by compressing natural language tokens into a small number of compressed tokens. When the number of compressed tokens is the same, the computational savings are nearly identical. Therefore, inference speed improvements are not analyzed in this paper.

\section{Related Work}
\label{related-work}

This work is related to prompt compression and LLM efficiency. There are two main approaches to reducing the number of prompt tokens: hard prompts and soft prompts.

Hard prompt methods identify and delete low-information content in the prompt. Li et al. proposed SelectiveSentence in 2023, which identifies rich-information content at the sentence or word level by calculating entropy or perplexity \cite{li-etal-2023-compressing}. Later, Jiang et al. proved that LLMs could understand incomplete words or sentences, leading to the development of LLMLingua and LongLLMLingua, which delete useless tokens even if fluency is interrupted \cite{jiang-etal-2023-llmlingua, jiang2023longllmlingua}.

Soft prompt methods compress natural language tokens into a small number of special tokens. Wingate et al. optimized the KL divergence between the answers generated by the original prompt and the compressed prompt, but this method lacked generalization, requiring training for each new prompt \cite{wingate-etal-2022-prompt}. Mu et al. solved this by proposing GIST tokens, but their limitations included the need for fine-tuning the original LLM and the short length of prompts to be compressed, typically less than thirty tokens \cite{mu2024learning}. ICAE solved these issues by pretraining the compression model and avoiding additional parameters during decoding, allowing compression of texts up to around 500 tokens without changing the original LLM \cite{ge2024incontext}. However, the maximum compression ratio of ICAE is about 15x. To increase the text length for compression, Chevalier proposed AutoCompressor, which recursively compresses the prompt but, like GIST tokens, is limited to fine-tuned LLMs and a complex training process \cite{chevalier2023adapting}. Other works explore text compression within paragraphs and soft prompt applications in agent domains \cite{ren2023context, jiang2024hierarchical}. Soft prompts are also applied in retrieval-augmented generation (RAG) through xRAG and COCOM \cite{cheng2024xrag, rau2024context}. Additionally, special soft prompts that do not retain detailed information but perform specific functions have been investigated. Todd et al. identified tokens for various functions such as generating antonyms or translating across languages \cite{todd2023function}, while Shen et al. used special tokens to enhance LLM performance in domain-specific tasks \cite{shen2024tag}.

The 500xCompressor proposed in this paper is a generalized soft prompt method capable of achieving high compression ratios up to 480x. It adapts the architecture of ICAE but utilizes K V values for the compressed tokens instead of their embeddings. This approach reduces mistakes, hallucinations, and information loss, leading to higher performance in regeneration and QA tasks.

\section{Conclusions}

This paper proposes 500xCompressor, a prompt compression method capable of compressing any text and all tokens within it. 500xCompressor achieves a high compression ratio while retaining most capabilities of non-compressed prompts. This method proves that current prompts are highly compressible, prompting further research into compression mechanisms and applications.

Future work would involve more extensive experiments and exploration of additional applications. Although 500xCompressor has shown good generalization ability on downstream tasks, it was pretrained and fine-tuned on relatively small corpora and QA datasets. Increasing the size and diversity of the training materials is expected to enable 500xCompressor to handle more tasks and achieve better results. Additionally, compressed texts could be applied in various contexts, including in-context learning, reasoning, personalized LLMs, RAG, and even role-playing. This method reduces response time and cost, enhancing user experience.

\section*{Ethics Statement}
This research did not involve any studies with human participants or animals performed by any of the authors. Therefore, no ethical approval was required for this study. All data and materials were collected in a manner consistent with ethical guidelines, ensuring no ethical concerns are present.

\section*{Availability Statement}
The code, datasets, models, and demo related to this study have been uploaded to the open source community at https://github.com/ZongqianLi/500xCompressor. For more information, additional data, or specific requests, please feel free to contact the first author or the corresponding author.

\bibliography{aaai25}

\begin{table*}[t]
    \centering
    \begin{tabular}{ccccccc}
        \hline
        {} & \multicolumn{3}{c}{\textbf{Pretraining}} & \multicolumn{3}{c}{\textbf{Finetuning}} \\
        {} & 500$\rightarrow$16 & 500$\rightarrow$4 & 500$\rightarrow$1 & 500$\rightarrow$16 & 500$\rightarrow$4 & 500$\rightarrow$1 \\
        \hline
        Total steps & 42000 & 66600 & 103800 & 20000 & 15000 & 10000 \\
        Warm-up steps & 300 & 300 & 300 & 300 & 300 & 300 \\
        Learning rate & 1e-4 & 1e-4 & 1e-4 & 5e-5 & 5e-5 & 5e-5 \\
        Batch size & 4 & 4 & 4 & 4 & 4 & 4 \\
        Optimizer & AdamW & AdamW & AdamW & AdamW & AdamW & AdamW \\
        \hline
    \end{tabular}
    \caption{Training parameters for 500xCompressor and ICAE.}
    \label{training-parameters}
\end{table*}

\begin{figure*}[t!]
\centering
\begin{subfigure}[b]{0.33\textwidth}
    \centering
    \includegraphics[width=\textwidth]{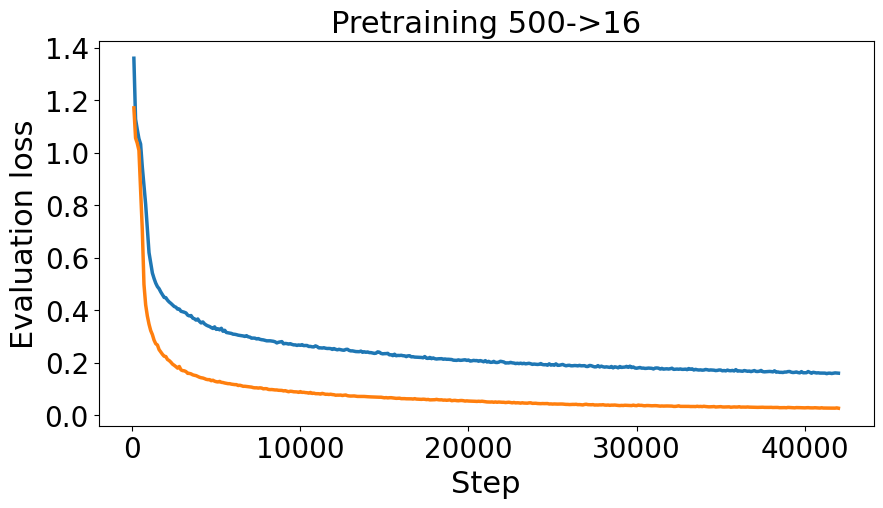}
    \label{pretrain-loss-1}
\end{subfigure}
\begin{subfigure}[b]{0.33\textwidth}
    \centering
    \includegraphics[width=\textwidth]{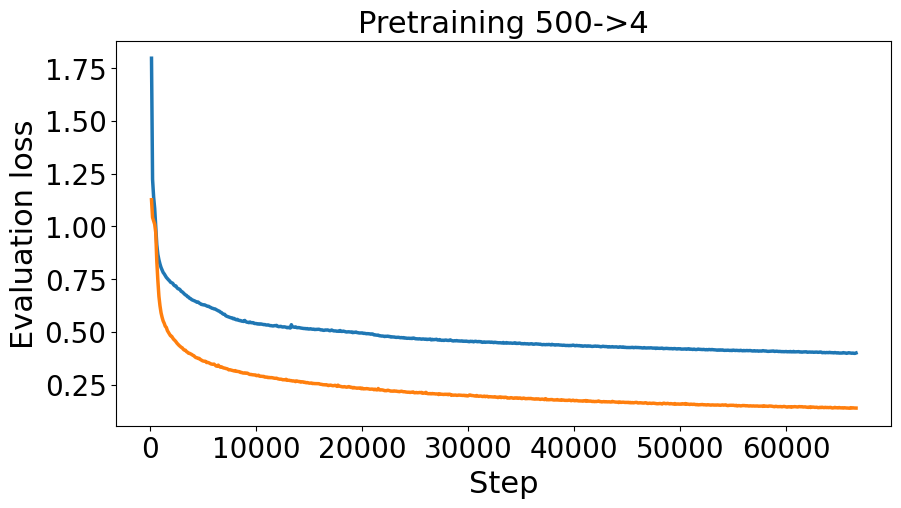}
    \label{pretrain-loss-2}
\end{subfigure}
\begin{subfigure}[b]{0.33\textwidth}
    \centering
    \includegraphics[width=\textwidth]{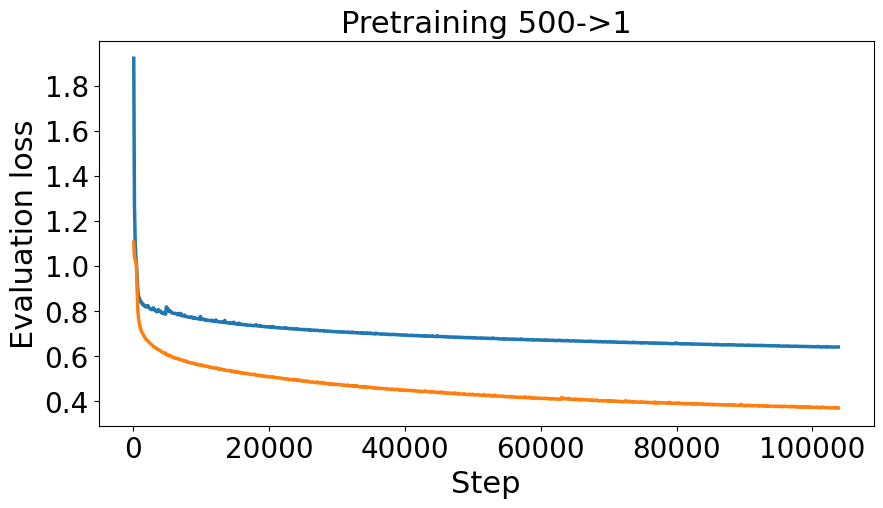}
    \label{pretrain-loss-3}
\end{subfigure}
\\
\begin{subfigure}[b]{0.33\textwidth}
    \centering
    \includegraphics[width=\textwidth]{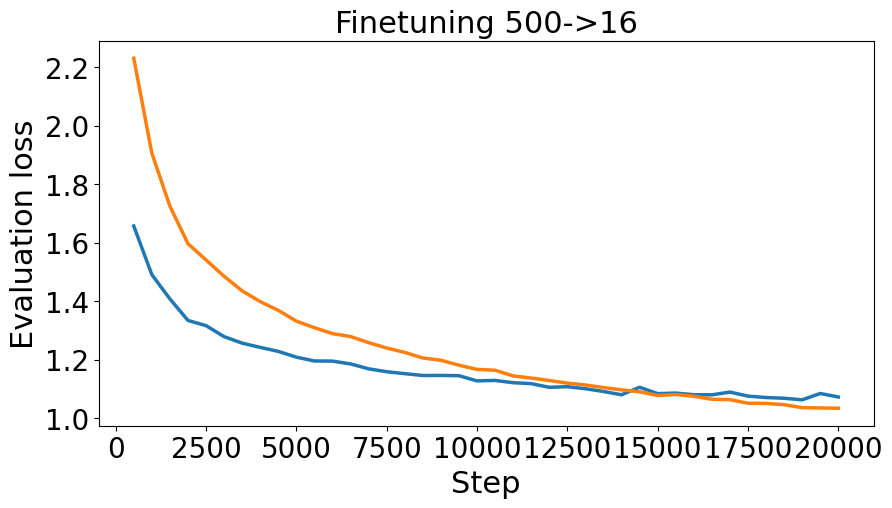}
    \label{fintune-loss-1}
\end{subfigure}
\begin{subfigure}[b]{0.33\textwidth}
    \centering
    \includegraphics[width=\textwidth]{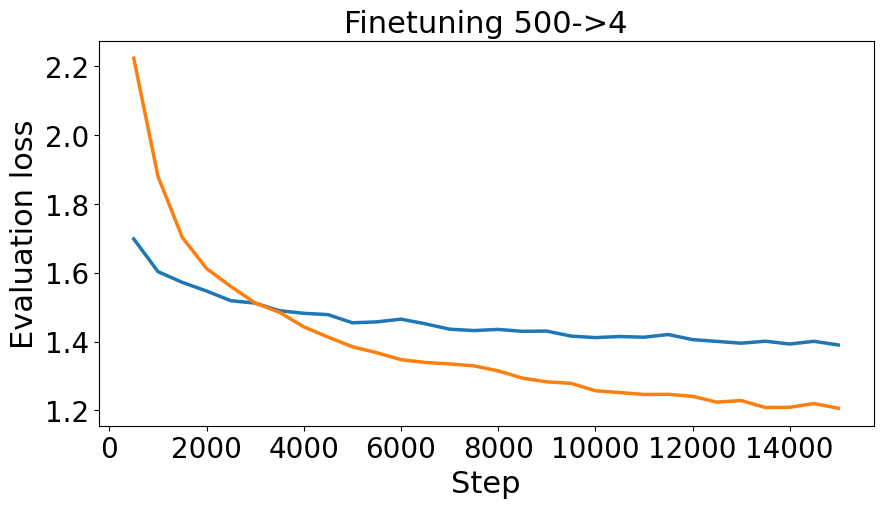}
    \label{fintune-loss-2}
\end{subfigure}
\begin{subfigure}[b]{0.33\textwidth}
    \centering
    \includegraphics[width=\textwidth]{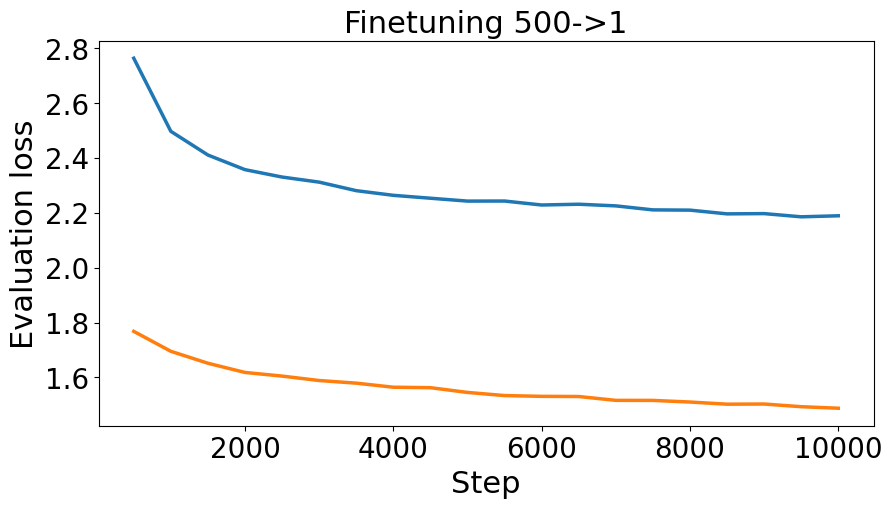}
    \label{fintune-loss-3}
\end{subfigure}
\\
\vspace{-1.5em}
\begin{subfigure}[b]{0.35\textwidth}
    \centering
    \includegraphics[width=\textwidth]{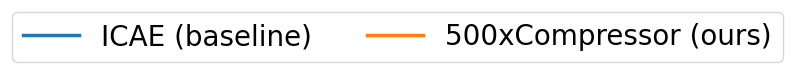}
    \label{loss-legend}
\end{subfigure}
\vspace{-1.5em}
\caption{Evaluation loss for 500xCompressor and ICAE during pretraining and fine-tuning.}
\label{loss}
\end{figure*}

\begin{table*}[ht!]
    \centering
    \begin{tabular}{ccccccc}
        \hline
        {} & \multicolumn{3}{c}{\textbf{Arxiv Corpus}} & \multicolumn{3}{c}{\textbf{ArxivQA Dataset}} \\
        {} & Train & Development & Test & Train & Development & Test \\
        \hline
        Number of data records & 2353924 & 3000 & 2500 & 250000 & 2500 & 5000 \\
        Knowledge cutoff & Pre 07/2023 & 01-04/2024 & 01-04/2024 & Pre 07/2023 & Pre 07/2023 & 01-04/2024 \\
        Source & \multicolumn{3}{c}{Abstracts pubslihed in Arxiv} & \multicolumn{2}{c}{Train set of Arxiv Corpus} & Test set of Arxiv Corpus \\
        \hline
    \end{tabular}
    \caption{Detailed information about the Arxiv Corpus and the ArxivQA dataset.}
    \label{corpus-qadataset}
\end{table*}

\begin{figure*}[t!]
\centering
\begin{subfigure}[b]{0.245\textwidth}
    \centering
    \includegraphics[width=\textwidth]{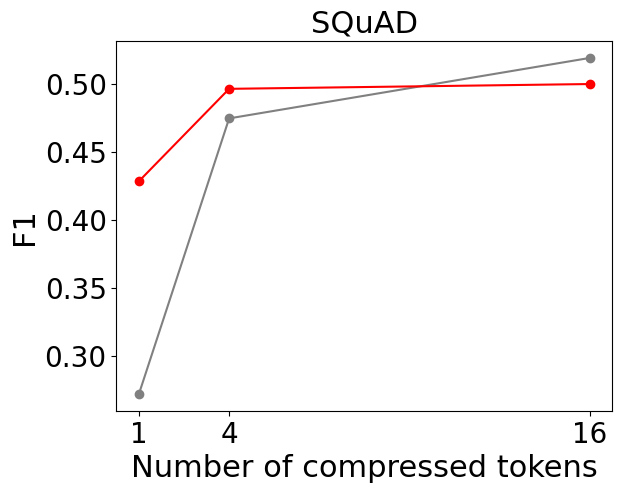}
    \label{squad-f1}
\end{subfigure}
\begin{subfigure}[b]{0.245\textwidth}
    \centering
    \includegraphics[width=\textwidth]{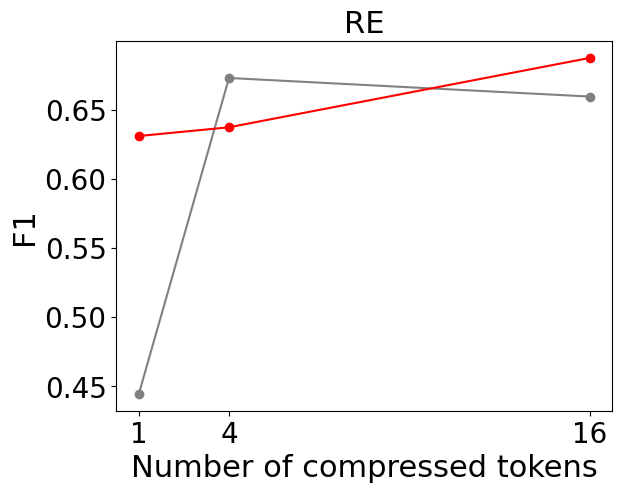}
    \label{re-f1}
\end{subfigure}
\begin{subfigure}[b]{0.245\textwidth}
    \centering
    \includegraphics[width=\textwidth]{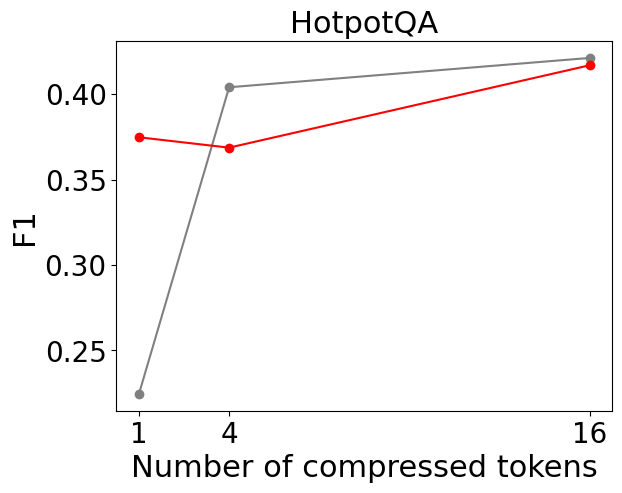}
    \label{hotpotqa-f1}
\end{subfigure}
\begin{subfigure}[b]{0.245\textwidth}
    \centering
    \includegraphics[width=\textwidth]{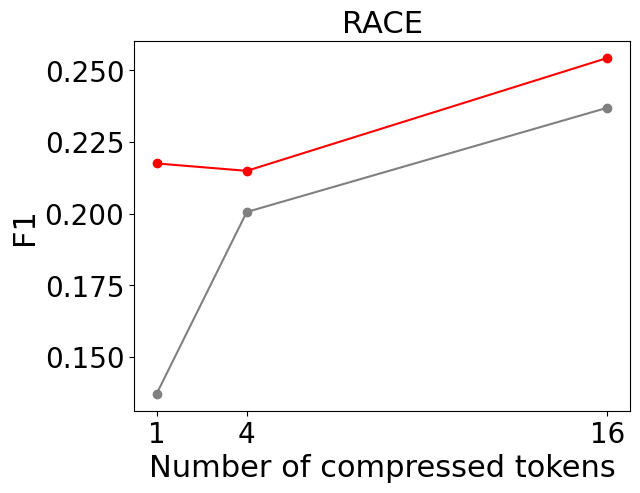}
    \label{race-f1}
\end{subfigure}
\\
\vspace{-1.5em}
\begin{subfigure}[b]{0.3\textwidth}
    \centering
    \includegraphics[width=\textwidth]{./Figures/reg-qa-legend.png}
    \label{benchmark-legend}
\end{subfigure}
\vspace{-1.5em}
\caption{Evaluation results for QA based on the compressed texts generated by 500xCompressor (ours) and ICAE (baseline).}
\label{benchmarks-figures}
\end{figure*}

\begin{table*}[ht!]
    \centering
    \begin{tabular}{ccccccccccccc}
        \hline
        {} & \multicolumn{2}{c}{\textbf{ArxivQA}} & \multicolumn{2}{c}{\textbf{SQuAD}} & \multicolumn{2}{c}{\textbf{RE}} & \multicolumn{2}{c}{\textbf{HotpotQA}} & \multicolumn{2}{c}{\textbf{RACE}} & \multicolumn{2}{c}{\textbf{Average}} \\
        {} & \multicolumn{4}{c}{Information Extraction} & \multicolumn{2}{c}{Relationships} & \multicolumn{2}{c}{Multi-hop} & \multicolumn{2}{c}{Comprehension} & \multicolumn{2}{c}{} \\
        \cline{2-13}
         & F1 & EM & F1 & EM & F1 & EM & F1 & EM & F1 & EM & F1 & EM \\
        \hline
        Zero-shot & 50.82 & 0.29 & 42.39 & 2.77 & 36.08 & 2.61 & 42.27 & 7.92 & 38.74 & 4.25 & 42.06 & 3.57 \\
        Instruct & 100 & 100 & 100 & 100 & 100 & 100 & 100 & 100 & 100 & 100 & 100 & 100 \\
        \hline
        Ours (500$\rightarrow$16) & 72.93 & 101 & 70.79 & 77.70 & 96.31 & 94.89 & 60.09 & 53.23 & 64.34 & 78.72 & 72.89 & 81.29 \\
        ICAE (500$\rightarrow$16) & 69.23 & 97.81 & 73.51 & 82.56 & 92.40 & 83.98 & 60.70 & 56.88 & 59.95 & 68.08 & 71.16 & 77.86 \\
        $\Delta$ & \textcolor{darkgreen}{3.70} & \textcolor{darkgreen}{4.07} & \textcolor{darkred}{-2.72} & \textcolor{darkred}{-4.86} & \textcolor{darkgreen}{3.91} & \textcolor{darkgreen}{10.91} & \textcolor{darkred}{-0.61} & \textcolor{darkred}{-3.64} & \textcolor{darkgreen}{4.39} & \textcolor{darkgreen}{10.63} & \textcolor{darkgreen}{1.73} & \textcolor{darkgreen}{3.42} \\
        \hline
        Ours (500$\rightarrow$4) & 68.51 & 77.43 & 70.29 & 74.87 & 89.29 & 85.96 & 53.13 & 45.58 & 54.38 & 56.38 & 67.12 & 68.04 \\
        ICAE (500$\rightarrow$4) & 59.58 & 54.87 & 67.21 & 71.99 & 94.28 & 95.78 & 58.23 & 55.00 & 50.75 & 48.93 & 66.01 & 65.32 \\
        $\Delta$ & \textcolor{darkgreen}{8.92} & \textcolor{darkgreen}{22.56} & \textcolor{darkgreen}{3.08} & \textcolor{darkgreen}{2.88} & \textcolor{darkred}{-4.98} & \textcolor{darkred}{-9.82} & \textcolor{darkred}{-5.09} & \textcolor{darkred}{-9.41} & \textcolor{darkgreen}{3.63} & \textcolor{darkgreen}{7.44} & \textcolor{darkgreen}{1.10} & \textcolor{darkgreen}{2.72} \\
        \hline
        Ours (500$\rightarrow$1) & 53.18 & 77.14 & 60.66 & 61.84 & 88.40 & 85.25 & 54.01 & 48.92 & 55.03 & 54.25 & 62.26 & 65.48 \\
        ICAE (500$\rightarrow$1) & 36.67 & 24.74 & 38.50 & 28.97 & 62.30 & 50.98 & 32.35 & 21.73 & 34.70 & 29.83 & 40.90 & 31.25 \\
        $\Delta$ & \textcolor{darkgreen}{16.51} & \textcolor{darkgreen}{52.40} & \textcolor{darkgreen}{22.16} & \textcolor{darkgreen}{32.86} & \textcolor{darkgreen}{26.10} & \textcolor{darkgreen}{34.26} & \textcolor{darkgreen}{21.65} & \textcolor{darkgreen}{27.19} & \textcolor{darkgreen}{20.33} & \textcolor{darkgreen}{24.42} & \textcolor{darkgreen}{21.35} & \textcolor{darkgreen}{34.23} \\
        \hline
    \end{tabular}
    \caption{Normalized evaluation results across five benchmarks focusing on various capabilities.}
    \label{benchmarks-normalized}
\end{table*}

\appendix

\section{Appendices}

\subsection{Model Training}

The training parameters for 500xCompressor and ICAE are detailed in Table \ref{training-parameters}. The evaluation losses for both 500xCompressor and ICAE are illustrated in Figure \ref{loss}.

\subsection{Arxiv Corpus and ArxivQA Dataset}

The detailed information for Arxiv Corpus and the ArxivQA dataset is shown in Table \ref{corpus-qadataset}.

The prompt to generate the QA pairs:

\begin{quote}
\begin{scriptsize}
\begin{lstlisting}[breaklines=true, basicstyle=\ttfamily, numbers=none, breakindent=0pt, xleftmargin=0pt, ]
context: {truncated_context}
task: design the {number} best extractive question answering pairs for the context to test information loss
requirement: the question should be direct; the question should try to use the same words in the context; the answer should directly appear in the context (a span of the context); the answer should not be in the question; just output the results in format and do not output other words
output json format: [{"id":1, "question": "", "answer": ""}, {"id":2, "question": "", "answer": ""}, ...]
\end{lstlisting}
\end{scriptsize}
\end{quote}

\subsection{Question Answering}

Prompt for QA benchmarks:

\begin{quote}
\begin{scriptsize}
\begin{lstlisting}[breaklines=true, basicstyle=\ttfamily, numbers=none, breakindent=0pt, xleftmargin=0pt, ]
Please finish the extractive question answering task. Just output the answer. Context: {context} Question: {question} Answer: 
\end{lstlisting}
\end{scriptsize}
\end{quote}

The evaluation results for SQuAD, RE, HotpotQA, and RACE are in Figure \ref{benchmarks-figures}. The normalized results for the benchmarks are in Table \ref{benchmarks-normalized}.

\end{document}